# Self-Prompting Small Language Models for Privacy-Sensitive Clinical Information Extraction

Yao-Shun Chuang[1*], Tushti Mody[1], Uday Pratap Singh[1], Shirindokht Shiraz[2], Chun-Teh Lee[3], Ryan Brandon[4], Muhammad F Walji[1], Xiaoqian Jiang[1], Bunmi Tokede[3*]

[1] McWilliams School of Biomedical Informatics, The University of Texas Health Science Center at Houston, Houston, TX 77030, United States,

[2] School of Public Health, The University of Texas Health Science Center at Houston, Houston, TX 77030, United States,

[3] School of Dentistry, The University of Texas Health Science Center at Houston, Houston, TX 77054, United States

[4] Willamette Dental and Skourtes Institute, Hillsboro, OR 97123, United States

[*] Corresponding author(s). E-mail(s): Yao-Shun.Chuang@uth.tmc.edu ; Oluwabunmi.Tokede@uth.tmc.edu

## Abstract

**Objective:**

Clinically meaningful information in dental progress notes remains largely embedded in unstructured free text, limiting its reuse for research and quality improvement. Because these notes contain protected health information, extraction must run locally rather than relying on external proprietary models. We developed and evaluated a self-contained framework enabling locally deployable small language models (SLMs) to perform clinical named-entity recognition (NER) in dental notes without transmitting clinical text outside the institution.

**Methods:**

Using 1,200 annotated progress notes (200 training, 1,000 held-out gold standard) spanning 19 entity types, each candidate SLM generated, verified, refined, and evaluated its own entity-specific prompts, with self-prompt performance serving as a label-free model-selection signal. Inference combined two-stage multi-prompt ensembling consisting of sentence-level Boolean screening followed by entity extraction with union-based merging. Selected models were further adapted using QLoRA-based supervised fine-tuning (SFT) followed by direct preference optimization (DPO) under local hardware constraints. Performance was assessed by entity-level precision, recall, and F1 using soft matching.

**Results:**

Self-prompt generation performance was broadly concordant with gold-standard inference, supporting its use as a task-specific screening signal, and predicted task performance more reliably than general-purpose leaderboard rankings. Prompts incorporating entity descriptions and examples improved extraction, while error feedback added only modest benefit. SFT increased recall but encouraged over-extraction, and DPO recalibrated this balance by improving precision while largely preserving recall gains. Qwen2.5-14B-Instruct with DPO achieved the best overall performance (0.864 micro F1, 0.837 macro F1), with gains across most models advanced to DPO.

**Conclusion:**

Self-generated prompt optimization combined with lightweight preference-based post-training supports accurate, privacy-preserving clinical information extraction from unstructured dental notes using locally deployed SLMs. Because the framework makes no dentistry-specific assumptions, it offers an extensible pathway for privacy-sensitive clinical NER more broadly, pending external validation.

## 1 Introduction

Electronic health records (EHRs) were expected to transform routine clinical documentation into real-world data for improving care quality, supporting research, and guiding more efficient use of healthcare resources[1,2]. These benefits have been constrained, however, by documentation burden, fragmented workflows, and the persistence of unstructured data[3–5]. The problem is particularly acute in dentistry, where EHRs are still often used primarily as digital versions of paper records, with limited secondary use beyond scheduling and billing[6–8]. Much of the clinically meaningful information in dental records, including diagnoses, social determinants of health, procedures, findings, treatment plans, and outcomes, remains embedded in free-text notes, leaving data collected during routine dental care underused for research, quality improvement, and learning health system development[7–10].

Extracting this information at scale is difficult. Dental notes express findings, procedures, and treatment plans through spatial, procedural, and abbreviation-heavy conventions that limit the transferability of natural language processing (NLP) methods developed for general medical text[11–16]. The deeper obstacle, however, is shared across clinical named entity recognition (NER). Documentation is unstructured and domain-specific, resisting methods built on cleaner corpora; more fundamentally, it is privacy-sensitive and cannot leave the institution, ruling out the common reliance on large external models and requiring extraction to run locally.

Large language models (LLMs) and emerging agentic frameworks have shown promise on language-intensive clinical tasks and offer a path toward structuring such heterogeneous documentation[17–24]. Yet frontier proprietary LLMs remain difficult to adopt clinically: clinical notes contain protected health information that constrains cloud-based processing, and the cost and infrastructure demands of large models further limit their use in settings such as dental schools and practice-based research networks[25–27]. Together, these constraints motivate the evaluation of locally deployable small language models (SLMs) for privacy-sensitive NER in dental clinical notes.

To address this need, we developed and evaluated a locally deployable SLM framework for dental clinical NER from unstructured progress notes. We define SLMs here as open-weight instruction-tuned models with fewer than roughly 30 billion parameters, small enough to deploy and adapt locally without dedicated large-scale infrastructure. This study makes three contributions. First, we introduce a self-generated instruction-prompting framework in which each candidate SLM generates, verifies, refines, and evaluates its own entity-specific prompts, reducing reliance on manual prompt engineering and keeping all clinical text within the institution. Second, we implement a two-stage, multi-prompt ensemble inference strategy combining sentence-level Boolean screening with entity extraction. Third, we demonstrate that QLoRA-based SFT followed by DPO can further adapt selected models under local hardware constraints. Although developed on dental notes, the framework makes no dentistry-specific assumptions and extends to other privacy-sensitive clinical NER settings.

## 2 Related Work

### 2.1 Clinical information extraction and NLP in dentistry

NLP has enabled large-scale extraction of clinically meaningful information from unstructured medical text, supporting applications such as phenotyping, surveillance, and quality measurement; reviews of clinical information extraction and of systems for standardizing unstructured clinical data document both the breadth of these applications and their reliance on domain adaptation[11–13]. Its application in dentistry, however, remains comparatively limited[16]. Dental notes pose distinct challenges because findings, procedures, and treatment plans are frequently expressed through tooth numbers, surfaces, quadrants, abbreviations, shorthand, and institution-specific documentation habits. These conventions constrain the direct transfer of methods developed for general medical text and call for approaches that accommodate the spatial, procedural, and highly variable nature of dental language[14–16]. Prior dental NLP efforts have made progress within narrow scopes, including deep-learning approaches that structure dental records to support clinical decision-making[14] and prior work on cross-institutional entity extraction using generative models and synthetic notes[10], but systematic reviews indicate that most existing systems remain rule- or dictionary-based and target a single task or a small set of entities[15,16]. Robust extraction across a broad, multi-entity annotation schema, of the kind required to render dental progress notes broadly reusable, remains largely unaddressed.

### 2.2 LLMs and agentic AI for clinical text

LLMs perform strongly across language-intensive clinical tasks—medical question answering, text summarization, and information extraction—in some cases approaching or exceeding expert performance[17–19]. Most relevant here, Agrawal et al. showed that LLMs can serve as few-shot clinical information extractors, recovering structured entities with minimal task-specific supervision[19]. A complementary line of work equips models with agentic capabilities such as task decomposition, tool use, planning, and iterative self-refinement, exemplified by ReAct[20], Toolformer[21], Reflexion[22], and Self-Refine[23]. These capabilities suit clinical extraction, where reliable performance requires applying entity definitions consistently across heterogeneous documentation. A recent review identifies dental documentation as a core use case for LLM-based text mining and generation[24].

However, these methods are typically built for large, cloud-hosted models and rely on external orchestration and repeated model calls; in agentic settings especially, the iterative refinement and multi-step tool use that drive performance can require many API calls per input, incurring substantial cost at scale and conflicting with the privacy and resource constraints of clinical deployment. Our framework adapts the underlying ideas of self-generation and iterative refinement to operate entirely within a single locally deployable SLM, so that prompt generation, verification, and refinement occur without transmitting clinical text externally and without per-call API expense.

### 2.3 Privacy constraints and local small language models

Despite their promise, frontier proprietary LLMs remain difficult to adopt in clinical settings because of privacy, governance, cost, and infrastructure constraints[25–27]. Clinical notes routinely contain protected health information, so external cloud-based processing depends on institutional review, contractual safeguards, and regulatory compliance, and several scoping reviews and analyses have examined the resulting privacy risks and mitigation strategies[28–30]. Proposed responses include secure institutional infrastructure for generative AI research[31] and privacy-preserving deployment that keeps clinical text within the local environment, including structured medical information retrieval with locally hosted models[32] and zero-egress, on-device deployment for sensitive domains such as mental health[33]. Local deployment mitigates these privacy concerns but introduces a practical tension, because large models impose substantial memory, hardware, and inference demands that may be prohibitive for dental schools, practice-based research networks, and other resource-constrained settings. This tension motivates the use of small language models, but prior privacy-preserving work has largely focused on deployment and retrieval rather than on how to select and adapt a local model for a specialized extraction task, which is the gap our framework addresses.

### 2.4 Model selection, prompt optimization, and adaptation

Selecting and adapting an open-weight model for clinical NER raises three connected challenges. The first is model selection. Public leaderboards based on benchmarks such as Instruction-Following Evaluation (IFEval) and Massive Multitask Language Understanding (MMLU) provide useful general signals but may not predict performance on specialized clinical extraction, because models differ in pretraining data, instruction tuning, reasoning behavior, context handling, and prompt sensitivity, and open clinical LLMs in particular have been shown to be sensitive to instruction phrasing[34–36]. In practice, researchers must therefore manually select candidate models and craft entity-specific prompts, repeating this effort whenever the model or target entity changes[37,38].

The second challenge is prompt construction. Automated prompt-optimization methods reduce this burden by enabling models to generate and refine task instructions, including APE[39], Self-Instruct[40], ProTeGi[41], and EvoPrompt[42], with broader developments surveyed elsewhere[37]. These methods can be effective but are often computationally or API-intensive and may yield prompts that are brittle across model architectures. A further limitation concerns the evaluation signal driving optimization: when candidate prompts are scored against model-generated instances or other self-referential criteria rather than independent gold annotations, the optimizer can improve the target metric without improving genuine extraction quality. Across these methods (Self-Instruct, EvoPrompt, APE, and ProTeGi[39–42]), prompt generation depends on an external proprietary model (GPT-3.5/4, InstructGPT, or davinci variants) and begins from human-written prompts or tasks, creating external dependencies and sending task content outside the local environment.

The third challenge is that prompting alone may be insufficient for robust clinical NER, motivating parameter-efficient post-training. QLoRA enables fine-tuning under reduced memory requirements, making adaptation feasible in resource-constrained clinical environments[43], and fine-tuning and prompt optimization have been shown to be complementary rather than mutually exclusive[44,45]. Supervised fine-tuning improves sensitivity for clinical and scientific information extraction[46,47], while batched inference amortizes the cost of shared instructions across queries[48]. Beyond supervised fine-tuning, direct preference optimization[49] refines model behavior using preference pairs, and applying it after supervised fine-tuning has been reported to benefit tasks requiring finer discrimination[46]. Individually, automated prompt optimization and parameter-efficient post-training are well established, but they are typically studied in isolation, and their combination has rarely been examined for clinical NER under local deployment constraints.

### 2.5 Synthesis

Prior work addresses the components of locally deployable clinical NER in isolation rather than as an integrated whole: dental NLP has relied on narrow rule- or dictionary-based systems; LLM and agentic methods offer flexible extraction but assume large, externally hosted models; privacy-preserving research has emphasized secure deployment and retrieval over task-specific model selection and adaptation; and automated prompt optimization and parameter-efficient post-training have been studied separately, each with known limitations in cost, brittleness, or robustness. No existing approach combines task-specific screening of locally deployable models, self-contained prompt generation that keeps clinical text within the institution, ensemble inference for robustness, and lightweight post-training into a single privacy-preserving pipeline. The present study addresses this gap.

## 3 Method

We developed a self-contained framework that enables a single locally deployable SLM to screen its own suitability for a target extraction task, generate and refine entity-specific prompts, perform ensemble inference, and undergo lightweight post-training, all within the institutional environment and without reliance on any external model. Figure 1 provides an overview. The framework proceeds in four stages. First, candidate models are screened using self-generated instruction prompts, with self-prompt performance serving as a task-specific selection signal in place of gold-standard labels. Second, each selected model generates, verifies, refines, and evaluates entity-specific prompts through an iterative optimization loop. Third, inference is performed using a two-stage, multi-prompt ensemble procedure combining sentence-level Boolean screening with entity extraction. Fourth, selected models are further adapted using QLoRA-based supervised fine-tuning followed by direct preference optimization. The remainder of this section describes the data source and annotation schema, each stage of the framework, and the evaluation metrics used throughout.

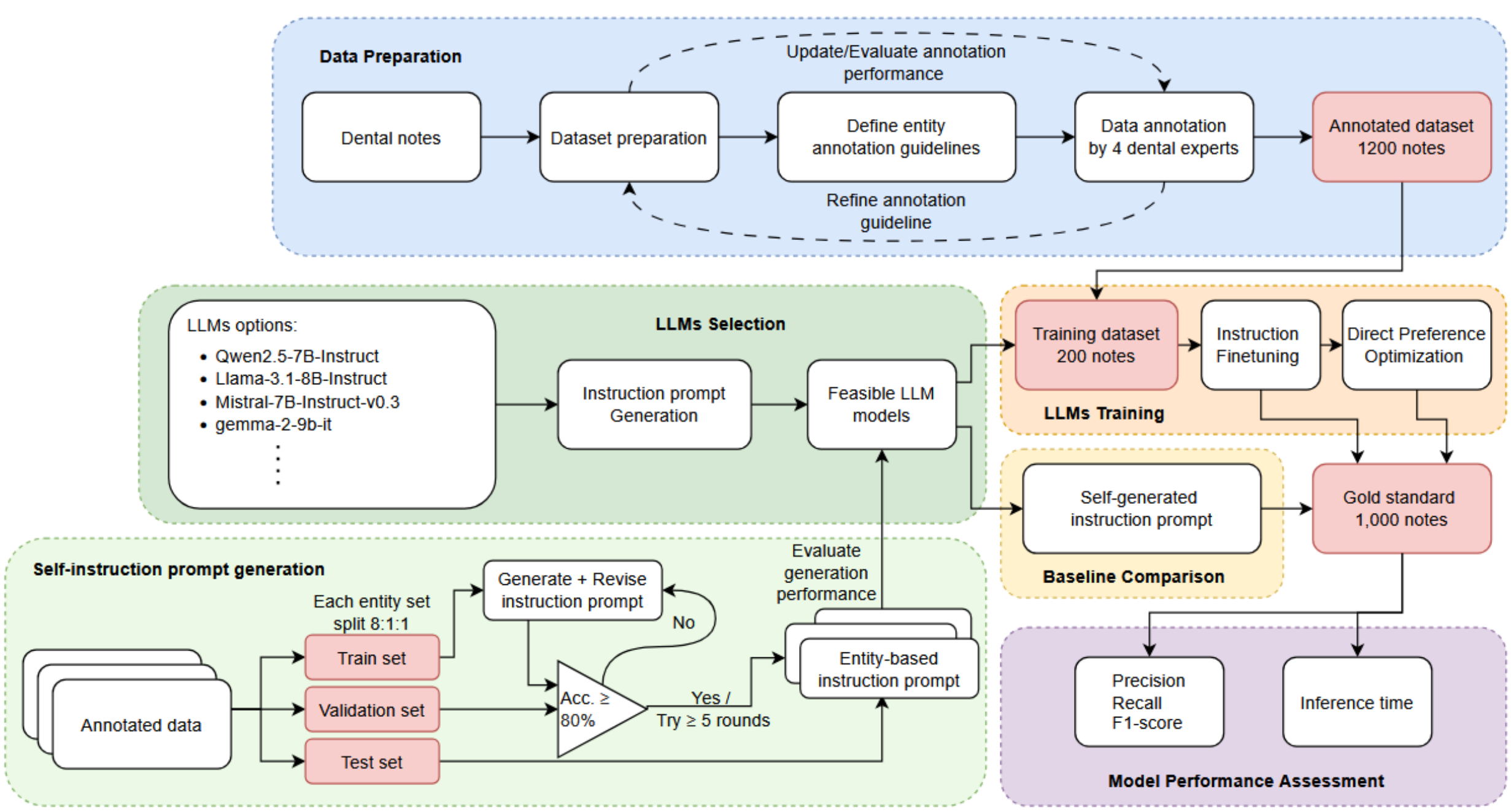

Figure 1. Overview of the study framework

## 3.1 Data source and annotation schema

The source dataset comprised 26,085 dental progress notes from 8,905 unique patients documented between January 1 and December 31, 2023. Notes were eligible if the patient was at least 16 years of age at the time of documentation, the note was a general progress note, and an examination or periodontal cleaning procedure was documented within ±1 day of the note date; orthodontic, laboratory, prescription, and treatment notes were excluded. From the eligible notes, 1,200 were randomly selected for this study and partitioned into a 200-note training set and a 1,000-note gold-standard evaluation set. The training set was used for all prompt generation, prompt optimization, SFT, and DPO preference-pair construction, whereas the 1,000-note gold-standard set was held out exclusively for final evaluation and was not used during prompt generation, prompt selection, SFT, DPO, model selection, or hyperparameter tuning.

Four dentists annotated the training set over four rounds, with group review after each round to refine the annotation schema and improve consistency. The gold-standard set was adjudicated in ten batches of 100 notes, with difficult or discordant cases escalated to a senior dentist for final resolution and adjudication decisions fed back to annotators to promote consistency across batches. Entity-level inter-annotator agreement based on soft matching is reported in Appendix A, Table 1. The annotation schema comprised 19 entity types; brief descriptions of each are provided in Appendix A, Table 2.

### 3.2 Self-generated instruction optimization

The core methodological contribution of this study is a self-contained, per-model prompt-optimization process in which each candidate SLM generates, verifies, refines, and evaluates its own entity-specific instruction prompts, without seed prompts and without reliance on any external model. This design directly addresses the privacy and portability constraints of clinical deployment: because the model that generates the prompts is also the model that performs extraction, no clinical text or representative example is transmitted outside the institutional environment, and prompts are tailored to each model rather than transferred across architectures whose pretraining and instruction-tuning differ.

Prior to instruction generation, dental notes were segmented into sentence-level units using spaCy with a RoBERTa-based pipeline, allowing each model to focus on a target sentence while limiting interference from surrounding content. For each target entity, sentences were divided into positive and negative groups according to whether they contained the entity of interest. An optional training-set revision step (Re) was then applied to the positive samples, in which the model deduplicated and curated near-duplicate or low-quality examples before prompt generation; if curation failed, the original samples were retained. Two further optional components governed prompt content: a brief manual description of the target entity (Desc) and inclusion of example cases (Ex). Descriptions were provided because, in domain-specific settings, examples alone may be insufficient for a model to capture the intended boundaries of an entity, whereas example use was made configurable because example content can be overly specific or otherwise undesirable in a final instruction. The complete prompts used for prompt generation and inference are provided in Appendix B.

Because the framework was designed for locally deployed SLMs, each candidate prompt underwent a verification step before evaluation. A candidate was accepted only if it satisfied the verification criteria covering clarity and self-containment, completeness, non-truncation, exclusion of embedded examples or references to example data, absence of redundant content, provision of inclusion and exclusion guidance for the target entity, and functional appropriateness for guiding extraction (Appendix A, Table 3). When the example-inclusion option (Ex) was enabled, the criteria prohibiting embedded examples or references to example data were waived, since example cases were intentionally permitted in the instruction under that configuration; the remaining criteria still applied. A candidate that failed verification entered a revision stage, in which the model revised the prompt according to seven revision guidelines addressing clarity and structure, explicitness of the task objective, redundancy removal, completeness, focus on the target entity, inclusion–exclusion and ambiguity guidance, and restraint in formatting (Appendix A, Table 4). The verified or revised prompt was then evaluated on the validation set using a two-phase chain-of-thought procedure consisting of Boolean screening followed by entity extraction, with performance assessed by entity-level precision, recall, and F1 score together with negative-sentence accuracy and positive-sentence soft-match accuracy. This cycle continued until the validation F1 score reached a predefined threshold (default 0.8) or a maximum of five rounds was completed; the five-round cap reflected the limited capacity of small models to further improve prompts for rare or difficult entities. An example of this iterative refinement across rounds is presented in Appendix A, Table 5. When the threshold was not reached within five rounds, the best-performing prompt on the validation set was retained. When the error-feedback option (Err) was enabled, incorrect validation examples were fed back into the subsequent round to guide revision. After optimization, the retained prompt was evaluated once on the held-out test set using the same two-phase procedure.

For entities with at least 10 positive samples, positive examples were split into training, validation, and test sets in an 80/10/10 ratio, and each subset was augmented with negative samples using multipliers of 3, 10, and 100, respectively. For entities with fewer than 10 positive samples, all positive samples together with negative training samples were used for instruction generation, and all positive samples together with negative test samples were used for evaluation.

Because of the relatively small size of the candidate models and the stochastic nature of their outputs, relying on a single generated instruction prompt was considered both unreliable and difficult to reproduce. Accordingly, each model generated 20 candidate instruction prompts per entity, and prompts were selected by F1 score on the prompt-generation evaluation. A threshold of 0.9 was applied: when more than three prompts exceeded the threshold, the three highest-scoring prompts were retained; when fewer than three exceeded it, all qualifying prompts were retained; and when none exceeded it, the three highest-scoring prompts were retained. Retaining multiple prompts rather than a single best prompt served two purposes: a single generated prompt may not be the highest-quality formulation available, and for difficult entities for which no prompt reached the threshold, retaining several complementary prompts allowed them to compensate for one another during ensemble inference.

Candidate models for this process were drawn from an initial screening based on publicly reported IFEval and MMLU performance; the resulting self-prompt-generation performance was then used as a task-specific criterion for identifying models suited to dental clinical NER, with screening results reported in Results. To assess whether this self-prompt screening signal tracked actual task performance, and how it compared with general-purpose leaderboard rankings, we quantified the rank concordance between each signal and gold-standard performance (Section 3.5). As noted in the Introduction, this procedure uses no dentistry-specific resources beyond the entity definitions and annotated examples supplied to it.

### 3.3 Two-stage, multi-prompt ensemble inference with chain-of-thought

For each clinical note and entity type, inference was performed using a two-stage, multi-prompt ensemble procedure. In Stage 1 (Boolean screening), each selected prompt independently determined whether a sentence contained the target entity, and a sentence advanced to extraction only if positive predictions exceeded half of the ensemble members. In Stage 2 (entity extraction), all selected prompts were applied to the screened sentences, and their outputs were merged by union-based deduplication while preserving first-seen order. Across all inference steps, greedy decoding was used to ensure deterministic output. This majority-vote screening followed by union extraction was intended to balance sensitivity and specificity: screening limited false-positive sentences, while union extraction recovered complementary mentions identified by different prompts.

The decision to combine multiple prompts rather than rely on a single instruction was motivated by two considerations. First, because prompt generation is stochastic, a single generated prompt may not be the highest-quality formulation; in development, single-prompt inference for Qwen2.5-7B-Instruct produced unstable performance across generation rounds on the gold-standard set, with micro F1 ranging from 0.534 to 0.657 and macro F1 from 0.541 to 0.601 (Appendix A, Table 6). Second, for difficult entities for which no prompt reached the selection threshold, retaining several complementary prompts allowed them to compensate for one another. We did not isolate the contribution of the ensemble relative to a single selected prompt in a controlled comparison; the variance observed across individual prompts motivated the design but does not by itself quantify the benefit of ensembling, which remains a target for future ablation.

To improve efficiency, inference used batched generation: inputs were grouped into token-aware batches constrained by the model context window and a configurable input ratio, and the framework automatically retried with a reduced token ratio when out-of-memory errors occurred. Global token usage was recorded throughout inference.

3.4 QLoRA-based supervised fine-tuning and preference optimization

Following prompt optimization, selected models were further adapted using a parameter-efficient, two-stage post-training procedure comprising SFT and DPO, both implemented with QLoRA to enable training under local hardware constraints. Both stages used 4-bit NF4 quantization, bfloat16 computation, double quantization, and low-rank adaptation; LoRA was configured with rank 16 applied to the q_proj, k_proj, v_proj, and o_proj modules. All post-training used only the 200-note training set; the 1,000-note gold-standard set was not used to construct SFT examples or DPO preference pairs and was reserved exclusively for final evaluation. Each model was trained for no more than two epochs under this shared configuration, and the checkpoint at which validation performance had stabilized was selected for final evaluation.

SFT was performed on chat-formatted instruction–response examples derived from the selected automatically generated prompts and the training-set annotations. Because several entity types were highly imbalanced at the sentence level, with some notes containing only a single sentence with the target entity (an approximate positive-to-negative ratio of 1:167), the SFT dataset was constructed with explicit handling of class imbalance. Positive sentences (those containing at least one target entity) were separated from negative sentences and divided into training and validation sets at a 9:1 ratio. Negative sentences were added to the training set at three times the number of positive sentences to support discrimination between positive and negative cases, whereas the validation set retained negative sentences at their natural frequency to provide a more realistic estimate of performance under imbalanced conditions.

DPO was subsequently applied to pairwise preference data consisting of a prompt, a chosen response, and a rejected response. The preference dataset was constructed entirely from incorrect SFT predictions: the chosen response corresponded to the original training-set annotation, and the rejected response was the erroneous output produced by the SFT model for the same input. Constructing preference pairs from a model's own outputs rather than from human-ranked comparisons is an established strategy for reducing annotation cost; accordingly, these pairs are best understood as annotation-guided optimization signals rather than conventional human preference data. The rationale for this stage was task-specific: SFT increased sensitivity to entity patterns and thereby improved recall, but at the cost of over-extraction and reduced precision, and DPO was used to recalibrate this balance by contrasting correct annotations against the model's own false-positive-prone outputs. This use of SFT to expand coverage followed by DPO to restore precision is consistent with prior clinical findings that DPO after SFT benefits tasks requiring finer discrimination[46], and we extend that observation to generative clinical NER by characterizing the specific precision–recall trade-off it addresses.

### 3.5 Evaluation metrics

Each entity type was evaluated using precision, recall, and F1 score. Precision reflects the proportion of predicted entities that were correct, recall reflects the proportion of gold-standard entities successfully identified, and F1 summarizes the balance between the two. A predicted entity was counted as correct when its similarity to the gold-standard annotation reached at least 80%, computed using the *partial_ratio* function in *RapidFuzz*. This soft-matching criterion was adopted to reduce the influence of trivial lexical or span-level differences between annotators while preserving substantive agreement. Macro-averaged precision, recall, and F1 were obtained by computing each metric separately per entity type and taking the unweighted mean across entity types. Micro-averaged metrics were obtained by pooling true positives, false positives, and false negatives across all entity types before computation, thereby weighting the result more heavily toward more frequent entities.

To evaluate self-prompt performance as a model-selection signal, we computed rank concordance between each model's self-evaluation metrics from the prompt-generation stage and its base gold-standard metrics using Kendall's $\tau$ (primary, for robustness at small $n$) and Spearman's $\rho$. The same correlations were computed against publicly reported IFEval and MMLU leaderboard scores for comparison. Self-prompt correlations used all 10 models that completed prompt generation; leaderboard correlations were restricted to the 8 with available scores. Correlations were computed separately for micro- and macro-averaged precision, recall, and F1, with p-values uncorrected for multiple comparisons and interpreted as descriptive.

## 4 Results

Entity distribution was highly imbalanced across the 19 annotation categories (Table 1). The training set contained 1,287 annotated entities and the gold-standard set 6,013, with higher counts for every entity type in the gold standard. Medication Taken and Systemic Condition were the most frequent entities in both sets, followed by Perio Diagnoses, Extent, Previous Medical Procedure, Stage, Grade, and the home-care behaviors (Brushing frequency, Flossing, Other Home Care). Race, Ethnicity, HbA1c Levels, and Family History Disease were comparatively rare. This pronounced long-tailed distribution motivated evaluating performance across both high- and low-frequency entities. Entity-level inter-annotator agreement is reported in Appendix A, Table 1.

Table 1. Entity distribution in the training set and gold standard.

| | | *Quantity* | | | | *Quantity* | |
|---|---|---|---|---|---|---|---|
| **No** | **Entity** | **Training set** | **Gold standard** | **No** | **Entity** | **Training set** | **Gold standard** |
| 1 | Age | 41 | 209 | 11 | HbA1c Levels | 20 | 59 |
| 2 | Race | 4 | 9 | 12 | Systemic Condition | 310 | 1,122 |
| 3 | Ethnicity | 13 | 39 | 13 | Family History Disease | 27 | 51 |
| 4 | Sex | 48 | 200 | 14 | Previous Medical Procedure | 109 | 422 |
| 5 | Perio Diagnoses | 48 | 500 | 15 | Medication Allergy | 40 | 164 |
| 6 | Stage | 25 | 366 | 16 | Medication Taken | 358 | 1,073 |
| 7 | Grade | 23 | 355 | 17 | Brushing frequency | 72 | 355 |
| 8 | Extent | 31 | 436 | 18 | Flossing | 33 | 241 |
| 9 | Subtype | 17 | 104 | 19 | Other Home Care | 29 | 205 |
| 10 | Social Factors | 39 | 103 | | *Total* | *1287* | *6013* |

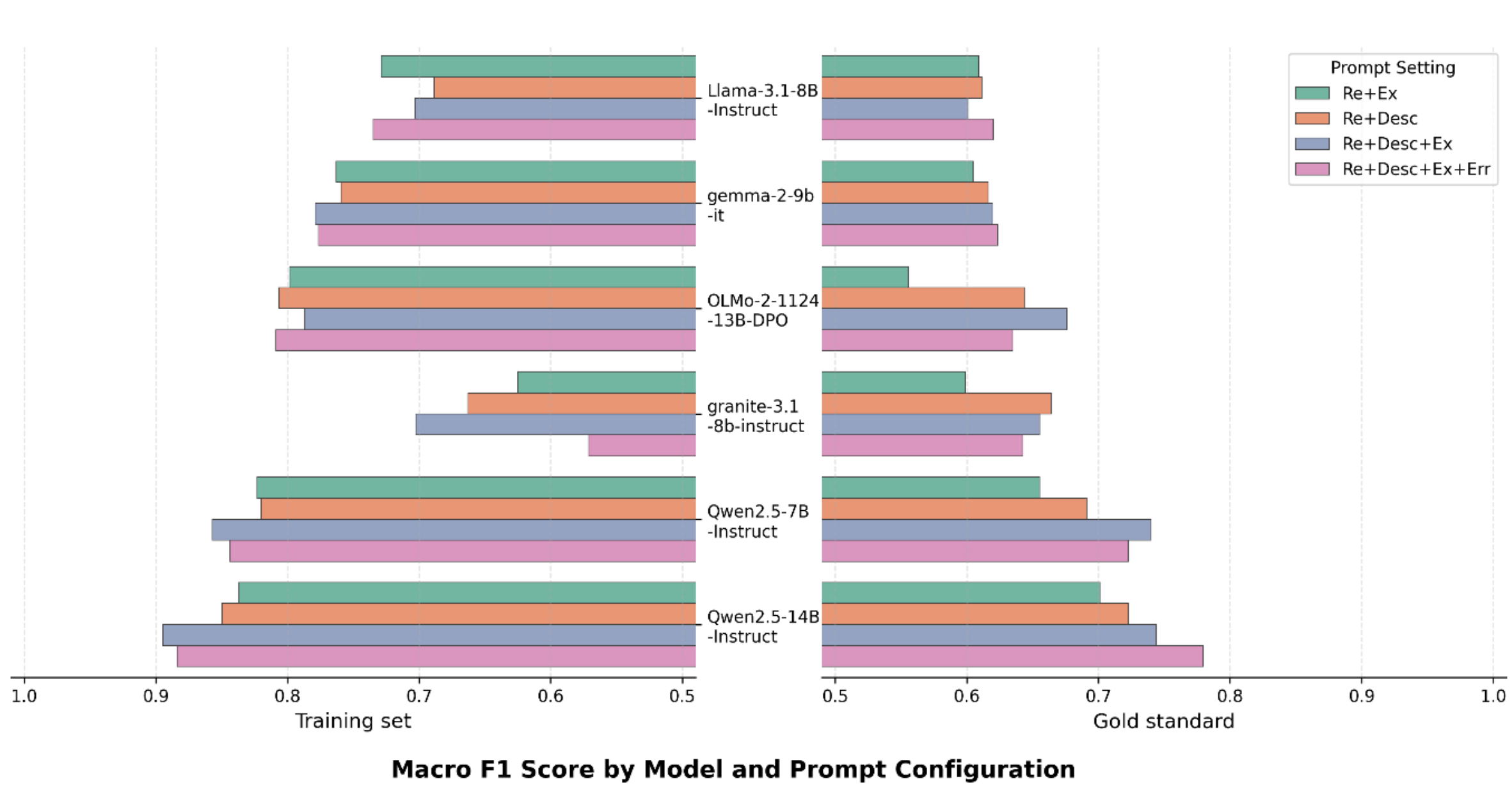

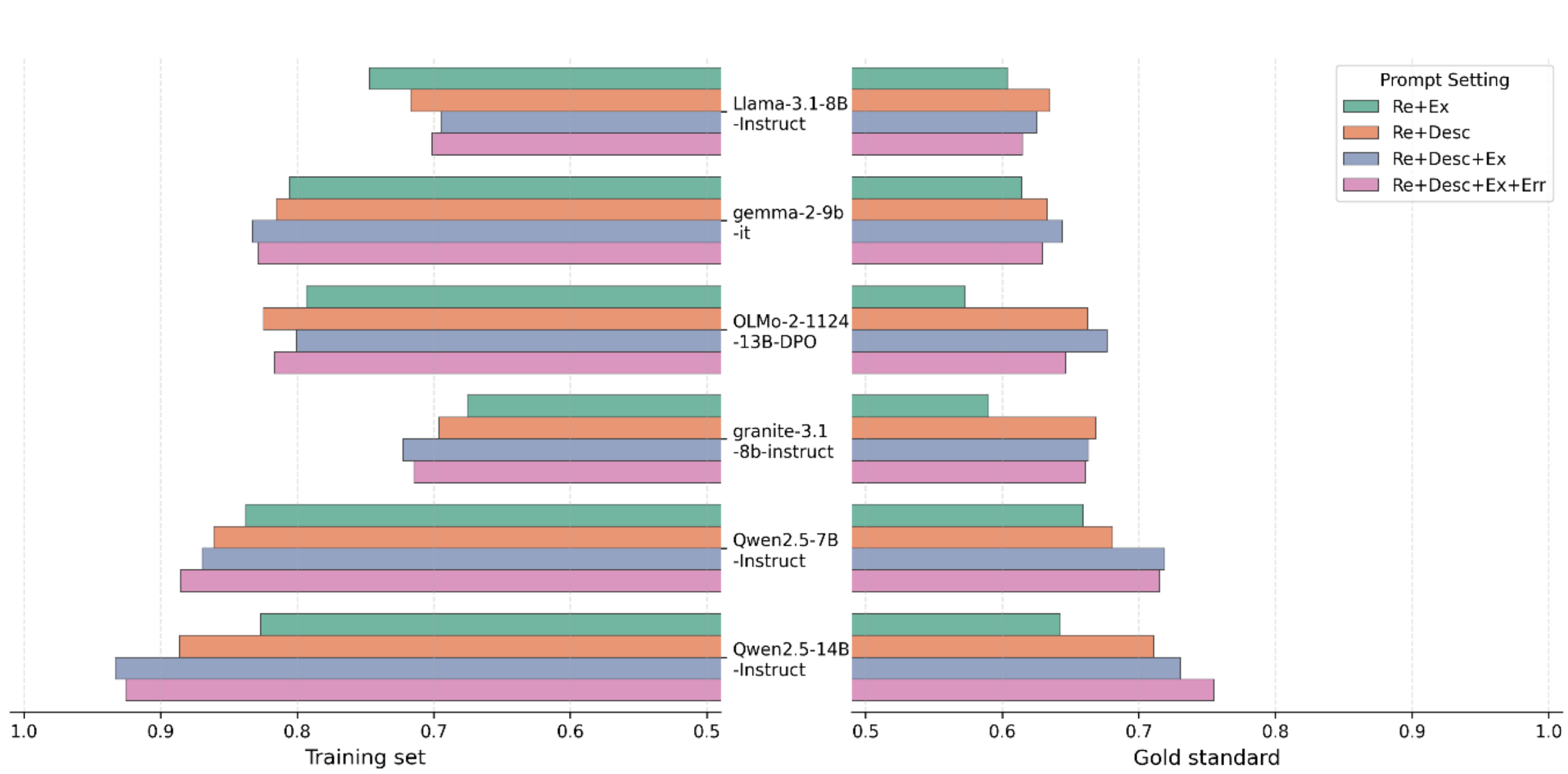

Figure 2. Prompt inference performance on the 200-note training set and 1000-note gold standard across prompt-generation configurations.

To reduce redundancy and bias from highly similar examples, training-set revision (Re) was applied for all models before prompt generation. We then evaluated four prompt-generation configurations combining Re with different prompt components:

representative examples (Re+Ex), entity descriptions (Re+Desc), descriptions plus examples (Re+Desc+Ex), and descriptions plus examples with error feedback during prompt evaluation (Re+Desc+Ex+Err) (Figure 2; full precision, recall, and F1 in Appendix A, Table 7). Across models, adding prompt components generally improved F1, though the magnitude and consistency varied by model family, with the largest gains for the Qwen models. On the gold-standard set, Qwen2.5-14B-Instruct achieved the best performance (0.780 micro F1, 0.755 macro F1) under Re+Desc+Ex+Err, followed by Qwen2.5-7B-Instruct (0.740 micro, 0.719 macro) under Re+Desc+Ex. Other families showed smaller, less consistent gains, indicating that richer prompts help but their benefit is model-dependent. Error feedback added only modest incremental benefit, suggesting most gains were attributable to entity descriptions and examples. Nonetheless, because Re+Desc+Ex+Err yielded the highest overall gold-standard performance and applying a single fixed configuration across models avoided per-model selection on held-out data, it was adopted as the standardized configuration for all subsequent comparisons.

Table 2. Self-prompt generation and self-evaluation across models

| Model | Time | Micro Average | | | Macro Average | | |
|---|---|---|---|---|---|---|---|
| | | P | R | F1 | P | R | F1 |
| Qwen2.5-7B-Instruct | 04:57:51 | 0.839 | 0.848 | 0.844 | 0.858 | 0.944 | 0.885 |
| Qwen2.5-14B-Instruct | 06:58:04 | **0.881** | 0.886 | **0.884** | **0.904** | **0.967** | **0.925** |
| Llama-3.1-8B-Instruct | 04:16:09 | 0.694 | 0.781 | 0.735 | 0.678 | 0.761 | 0.701 |
| Llama-3.2-3B-Instruct | 05:08:17 | 0.370 | 0.200 | 0.260 | 0.377 | 0.304 | 0.276 |
| Mistral-7B-Instruct-v0.3 | 09:55:37 | 0.430 | 0.342 | 0.381 | 0.375 | 0.314 | 0.277 |
| Mistral-Small-Instruct-2409 | 16:17:58 | 0.591 | 0.323 | 0.417 | 0.445 | 0.359 | 0.362 |
| OLMo-2-1124-13B-DPO | 11:16:17 | 0.753 | 0.873 | 0.809 | 0.812 | 0.862 | 0.817 |
| OLMo-2-1124-7B-DPO | 10:25:47 | 0.680 | 0.430 | 0.527 | 0.580 | 0.439 | 0.482 |
| gemma-2-9b-it | 04:54:06 | 0.661 | **0.941** | 0.776 | 0.756 | 0.957 | 0.829 |
| gemma-3-12b-it | 2 days 03:15:02 | - | - | - | - | - | - |
| granite-3.1-8b-instruct | 06:43:14 | 0.737 | 0.466 | 0.571 | 0.762 | 0.736 | 0.714 |
| Falcon3-Mamba-7B-Instruct | 4 days 09:24:04 | - | - | - | - | - | - |

*Abbreviation:* P, precision; R, recall; F1, F1 score.

Using IFEval and MMLU leaderboard performance as an initial screening criterion, we selected 12 candidate models for local evaluation of self-generated instructions across the 19 target entities. Table 2 reports self-prompt generation and self-evaluation performance, measured on each model's own held-out test split during prompt optimization rather than on the gold standard. Performance varied substantially across models. Among those with complete results, Qwen2.5-14B-Instruct achieved the highest micro- and macro-average F1 (0.884 and 0.925), followed by Qwen2.5-7B-Instruct (0.844 and 0.885); OLMo-2-1124-13B-DPO and gemma-2-9b-it also performed strongly, though gemma-2-9b-it showed high recall but comparatively low precision. Llama-3.2-3B-Instruct and the Mistral variants showed substantially lower, less balanced F1. Notably, this self-evaluation ranking broadly tracked subsequent gold-standard performance (Table 3), supporting the use of self-prompt performance as a task-specific screening signal. Runtime also differed markedly: gemma-3-12b-it and Falcon3-Mamba-7B-Instruct required prohibitively long runtimes, returned near-zero scores in generation logs, and were terminated and excluded from further comparisons.

After excluding the two models that did not complete prompt generation, we evaluated the remaining 10 locally deployable SLMs on the 1,000-note gold-standard set under three settings: the original instruction-tuned base model, QLoRA-based SFT, and DPO following SFT (Table 3). This comparison was designed to assess whether lightweight post-training could improve extraction performance beyond self-generated prompt-based inference alone.

Among the base models, evaluated with self-generated prompts under the standardized Re+Desc+Ex+Err configuration (Figure 2) and without fine-tuning, performance varied substantially (Table 3). Qwen2.5-14B-Instruct achieved the best overall performance, with the highest micro- and macro-average F1 (0.780 and 0.755), followed by Qwen2.5-7B-Instruct (0.723 and 0.715). granite-3.1-8b-instruct, OLMo-2-1124-13B-DPO, gemma-2-9b-it, and Llama-3.1-8B-Instruct showed intermediate performance, whereas Llama-3.2-3B-Instruct, OLMo-2-1124-7B-DPO, and the Mistral models performed less favorably. Notably, gemma-2-9b-it and Llama-3.1-8B-Instruct showed relatively high recall but lower precision, indicating a tendency toward overprediction. Overall, the Qwen family, particularly Qwen2.5-14B-Instruct, provided the strongest and most balanced baseline performance.

Table 3. Performance of instruction-tuned SLMs across base, SFT, and DPO settings.

| Models | Base | | | | | | SFT | | | | | | DPO | | | | | |
|---|---|---|---|---|---|---|---|---|---|---|---|---|---|---|---|---|---|---|
| | Micro Average | | | Macro Average | | | Micro Average | | | Macro Average | | | Micro Average | | | Macro Average | | |
| | P | R | F1 | P | R | F1 | P | R | F1 | P | R | F1 | P | R | F1 | P | R | F1 |
| Qwen2.5-7B-Instruct | 0.736 | 0.710 | 0.723 | 0.742 | 0.719 | 0.715 | 0.571 | 0.823 | 0.674 | 0.610 | 0.833 | 0.669 | 0.925 | 0.705 | 0.800 | 0.908 | 0.683 | 0.769 |
| Qwen2.5-14B-Instruct | 0.798 | 0.762 | ***<u>0.780</u>*** | 0.791 | 0.748 | ***<u>0.755</u>*** | 0.616 | 0.895 | 0.730 | 0.623 | 0.881 | 0.709 | 0.948 | 0.794 | ***<u>0.864</u>*** | 0.948 | 0.769 | ***<u>0.837</u>*** |
| Llama-3.1-8B-Instruct | 0.523 | 0.761 | 0.620 | 0.563 | 0.783 | 0.615 | 0.694 | 0.838 | ***<u>0.759</u>*** | 0.693 | 0.853 | ***<u>0.738</u>*** | 0.948 | 0.701 | 0.806 | 0.949 | 0.696 | 0.797 |
| Llama-3.2-3B-Instruct | 0.251 | 0.466 | 0.326 | 0.361 | 0.467 | 0.353 | 0.449 | 0.592 | 0.511 | 0.422 | 0.513 | 0.424 | | | | | | |
| Mistral-7B-Instruct-v0.3 | 0.296 | 0.378 | 0.332 | 0.361 | 0.296 | 0.267 | 0.158 | 0.868 | 0.267 | 0.216 | 0.846 | 0.307 | | | | | | |
| Mistral-Small-Instruct-2409 | 0.318 | 0.430 | 0.365 | 0.448 | 0.458 | 0.370 | 0.405 | 0.628 | 0.492 | 0.424 | 0.717 | 0.470 | | | | | | |
| OLMo-2-1124-13B-DPO | 0.565 | 0.725 | 0.635 | 0.652 | 0.696 | 0.646 | 0.433 | 0.863 | 0.577 | 0.463 | 0.842 | 0.572 | | | | | | |
| OLMo-2-1124-7B-DPO | 0.617 | 0.329 | 0.429 | 0.546 | 0.286 | 0.351 | 0.631 | 0.706 | 0.666 | 0.574 | 0.583 | 0.554 | 0.799 | 0.646 | 0.714 | 0.788 | 0.584 | 0.646 |
| gemma-2-9b-it | 0.510 | 0.803 | 0.624 | 0.588 | 0.759 | 0.630 | 0.575 | 0.637 | 0.604 | 0.650 | 0.649 | 0.622 | 0.910 | 0.010 | 0.020 | 0.626 | 0.021 | 0.035 |
| granite-3.1-8b-instruct | 0.646 | 0.639 | 0.643 | 0.690 | 0.663 | 0.661 | 0.382 | 0.892 | 0.535 | 0.439 | 0.892 | 0.547 | | | | | | |

*Note:* DPO was applied only to models with micro- or macro-average F1 above 0.6 after SFT; blank cells indicate models not advanced to DPO. P, precision; R, recall; F1, F1 score; SFT, supervised fine-tuning; DPO, direct preference optimization.

Post-training effects varied across models. Because several SFT-adapted models showed limited performance, with F1 below a practically useful level, DPO was applied only to models reaching micro- or macro-average F1 above 0.6 after SFT, focusing preference optimization on models with sufficient task competence. Among all configurations, Qwen2.5-14B-Instruct with DPO performed best, reaching micro- and macro-average F1 of 0.864 and 0.837 and improving over both its base and SFT settings. Qwen2.5-7B-Instruct and Llama-3.1-8B-Instruct also gained clearly after DPO, with final micro-average F1 of 0.800 and 0.806 and macro-average F1 of 0.769 and 0.797. OLMo-2-1124-7B-DPO improved most across stages, rising from 0.429/0.351 at baseline to 0.666/0.554 after SFT and 0.714/0.646 after DPO. Entity-level performance for the two best models, Qwen2.5-14B-Instruct (DPO) and Llama-3.1-8B-Instruct (DPO), is reported in Appendix A, Table 8.

The effect of SFT alone was less consistent. It improved several weaker baseline models, including Llama-3.2-3B-Instruct, Mistral-Small-Instruct-2409, and OLMo-2-1124-7B-DPO, but reduced performance for the Qwen models, OLMo-2-1124-13B-DPO, gemma-2-9b-it, and granite-3.1-8b-instruct. Preference optimization was also not uniformly beneficial: gemma-2-9b-it degraded markedly after DPO, with micro- and macro-average F1 falling to 0.020 and 0.035. Overall, post-training gains among models advanced to DPO were generally substantial, with gemma-2-9b-it the notable exception, whereas SFT alone produced more variable effects. Appendix A, Figure 1 shows performance across the baseline, SFT, and DPO stages for five SLMs; gold-standard inference times are given in Appendix A, Table 9.

Self-prompt performance was strongly concordant with gold-standard performance across the 10 models that completed prompt generation in Appendix A, Table 10. Concordance was highest for the F1 metrics used in model selection (micro-F1 Kendall's $\tau = 0.82$, Spearman's $\rho = 0.94$; macro-F1 $\tau = 0.85$, $\rho = 0.95$; all $p < 0.001$) and similarly high for precision (micro-precision $\tau = 0.82$, $\rho = 0.94$). Recall was less strongly predicted (micro-recall $\tau = 0.58$, macro-recall $\tau = 0.60$), though both remained significant. The agreement between Kendall's $\tau$ and Spearman's $\rho$ across all metrics indicates that the concordance was robust to the choice of rank-correlation coefficient. Because models were selected on F1, this rank agreement supports the use of self-prompt performance as a label-free screening signal.

The general-purpose leaderboard benchmarks were also positively associated with gold-standard performance, but more weakly and less consistently. Leaderboard scores were available for 8 of the 10 models; the two OLMo-2-1124 variants were excluded because the leaderboard reports only their base versions, whereas we evaluated the slightly stronger official DPO releases, making the published scores non-comparable. Among the 8 models with available leaderboard scores, MMLU showed moderate concordance with gold-standard F1 (micro-F1 $\tau = 0.64$, macro-F1 $\tau = 0.62$; both $p < 0.05$), while IFEval was weaker and did not reach significance (micro-F1 $\tau = 0.57$, macro-F1 $\tau = 0.55$; both $p \approx 0.06$). For both benchmarks, recall was poorly predicted and non-significant. Self-prompt performance therefore predicted task performance more strongly and more consistently than either benchmark, supporting task-specific self-evaluation over general-purpose leaderboard rankings when selecting locally deployable models for clinical NER, while indicating that leaderboard scores retain some, more limited, predictive value.

## 5 Discussion

This study developed and evaluated a self-generated instruction-prompt framework that enables locally deployable SLMs to perform clinical NER in unstructured dental progress notes, addressing a recurring obstacle: identifying models that achieve reliable task-specific performance under local hardware constraints while reducing reliance on manual prompt engineering. Because each candidate SLM generated and refined its own prompts using only local resources, the design avoids the prompt brittleness, model dependency, high computational or API costs, and susceptibility to self-induced reward hacking that constrain prior autonomous prompt-generation methods. By standardizing this process across 19 dental entity types, the framework offers a scalable approach for screening locally deployable models for domain-specific clinical NER, and extends to privacy-sensitive clinical NER more broadly.

A central finding is that self-prompt generation performance was broadly concordant with gold-standard inference (Tables 2 and 3), suggesting it can serve as a task-specific screening signal that ranks and selects candidate models without gold-standard labels or any external model. Such label-free, self-contained selection is particularly valuable in privacy-constrained settings, where annotated gold standards are costly and clinical text cannot leave the institution. General-purpose leaderboards predicted task performance more weakly: MMLU showed only moderate concordance and IFEval weaker, non-significant concordance, reinforcing the value of task-specific assessment over general leaderboards when adopting SLMs for specialized clinical extraction.

Richer self-generated prompts that incorporated entity descriptions and examples improved extraction performance, whereas additional error-feedback refinement provided only modest incremental benefit. This pattern suggests that concise entity-specific descriptions help guide SLMs toward the intended clinical meaning, particularly when annotation boundaries are subtle or when examples alone are insufficient to convey them. This comparison isolated the effect of the prompt-generation configuration: two-stage Boolean screening, chain-of-thought inference, multi-prompt ensembling, majority voting, and batching were held constant across configurations. The observed differences should therefore be attributed to prompt-generation design rather than to these fixed inference components, whose independent contributions remain to be quantified through further ablation.

Robustness was further supported by ensemble inference, which combined multiple optimized prompts through Boolean screening, entity extraction, and majority voting rather than relying on a single instruction. This reduced dependence on any individual prompt and mitigated the variability inherent to stochastic generation: screening limited false-positive sentences, while union-based extraction recovered complementary mentions identified by different prompts, balancing sensitivity and specificity and improving reproducibility across heterogeneous note structures. Together with self-generated prompt optimization, ensemble inference offers a practical strategy for adapting locally deployed SLMs to domain-specific clinical information extraction.

Lightweight post-training further improved small-model performance while preserving local feasibility[46,47]. QLoRA enabled parameter-efficient adaptation under reduced memory requirements[43], and token-aware batched generation reduced inference time substantially without degrading extraction, consistent with prior reports that batch prompting amortizes the cost of shared instructions across queries[48,50]. Within this regime, SFT and DPO played complementary roles: SFT increased sensitivity to entity patterns, reducing false negatives and improving recall[47], but the same adaptation encouraged over-extraction, lowering precision. DPO recalibrated this balance by contrasting correct annotations against the model's own false-positive-prone outputs[49], improving precision while largely preserving SFT's recall gains. Because rejected responses were generated automatically from incorrect SFT predictions rather than human-ranked, these pairs are best understood as annotation-guided optimization signals rather than conventional human preference data. The marked degradation of gemma-2-9b-it after DPO indicates that preference optimization is not uniformly beneficial across architectures and may destabilize performance when the model, preference data, or training configuration is poorly aligned.

Error analysis of the two best-performing models, Qwen2.5-14B-Instruct with DPO and Llama-3.1-8B-Instruct with DPO, showed that most residual errors arose in cases requiring contextual interpretation rather than recognition of explicit entity mentions. Both models reliably identified clearly stated entities but faltered when information was conveyed through abbreviations, irregular formatting, fragmented sentence structures, or modifier-heavy diagnostic statements. Sentence-level extraction compounded this difficulty, because isolated phrases or line breaks sometimes removed the contextual cues needed to distinguish patient demographics, smoking history, oral hygiene behaviors, laboratory values, medication status, and periodontal diagnoses. Errors also clustered at the boundaries of semantically related categories, such as existing systemic conditions versus family history of disease, medication allergy versus current medication, and previous medical versus dental or orthodontic procedures, and the models sometimes failed to separate patient-reported behaviors from clinician recommendations, negated statements, or irrelevant mentions. These patterns indicate that the principal remaining limitations of locally deployed SLMs lie in clinical ambiguity, fragmented documentation, entity-boundary overlap, and limited fine-grained contextual reasoning.

Several limitations should be acknowledged. First, the framework was evaluated using data from a single institution, so performance may not transfer directly to other settings. The framework itself introduces no dependence on this particular setting, which suggests it should extend to other privacy-sensitive clinical NER tasks; external validation across multiple institutions and clinical domains is nonetheless a priority for future work to test this directly. Second, because dental notes are highly unstructured, sentence-level preprocessing may have fragmented semantically related information into overly short or incomplete units, and the error analysis suggests this contributed to context-dependent failures. Future work should explore post-processing that merges adjacent short sentences or otherwise preserves broader context. Third, prompt generation was capped at 20 candidates per entity with the three best retained for inference; given the stochastic nature of LLM-based generation, evaluating additional candidates may yield further gains, at the cost of greater computational time. Fourth, all models were trained under a single shared hyperparameter and QLoRA configuration, and model-specific tuning may improve performance but was beyond the scope of this study. Finally, although prompt-generation configurations and post-training strategies were each evaluated separately, we did not perform a full factorial ablation of the fixed inference components. In particular, the independent contributions of two-stage Boolean screening, chain-of-thought inference, multi-prompt ensembling, and token-aware batching were not isolated; the variability observed across individual prompts motivated the ensemble design but does not by itself quantify its benefit. Systematically disentangling these components is an important direction for future work.

## 6 Conclusion

This study demonstrates that automated, self-generated prompt optimization combined with lightweight preference-based post-training can support accurate clinical information extraction from unstructured dental documentation using locally deployed SLMs. By having each candidate model generate, verify, refine, and evaluate its own entity-specific prompts, the framework requires no external model and no transmission of clinical text beyond the institution, and its self-prompt performance proved broadly concordant with gold-standard inference, offering a label-free signal for screening models suited to a target clinical task. Ensemble inference improved robustness to prompt-level variability, while QLoRA-based supervised fine-tuning followed by direct preference optimization recalibrated the precision-recall balance under realistic local hardware constraints. Performance varied substantially across models, and self-prompt performance predicted task performance more reliably than general-purpose leaderboard rankings, underscoring the value of task-specific assessment when selecting locally deployable models; Qwen2.5-14B-Instruct after DPO achieved the strongest overall results. Because the framework makes no assumptions specific to dentistry, it offers a practical and extensible pathway toward scalable, privacy-preserving, and resource-conscious clinical information extraction, pending external validation across additional institutions and clinical domains.

## Ethics approval and consent to participate

This study was reviewed by the Institutional Review Board at The University of Texas Health Science Center at Houston under protocol HSC-DB-21-0616. A waiver of patient informed consent was granted.

## Data availability

The study dataset is not currently publicly available because the research team are completing the required de-identification procedures and related legal review. Once these processes are finalized, the de-identified dataset will be made publicly available through an appropriate open-access repository.

## Code availability

The framework, including the source code and prompts used in this study, is publicly available on GitHub at: https://github.com/lifestrugglee/fullmouth.

## Acknowledgements

This work was supported by the National Institute of Dental and Craniofacial Research of the NIH under award number R56DE034086.

## Author contributions

Y.S.C. developed the methodology and drafted the manuscript. T.M., U.P.S., S.S., C.T.L, R.B. curated the data. M.W., X.J. and B.T. supervised the study and acquired funding. All authors reviewed and edited the manuscript and approved the final version.

## Competing interests

The authors declare no competing interests.

## Figure legends

Figure 1. Overview of the study framework

Figure 2. Prompt inference performance on the 200-note training set and 1000-note gold standard across prompt-generation configurations.

## Appendix A

Table 1. Soft entity-level inter-annotator agreement during training-set annotation.

| Round | Num. notes | Inter-Annotator Agreement |
|---|---|---|
| 1 | 50 | - |
| 2 | 50 | 61.8% |
| 3 | 50 | 77.5% |
| 4 | 50 | 85.3% |
| Total | 200 | |

*Note: Agreement was calculated using entity-level soft matching. Round 1 was used for calibration, allowing annotators to become familiar with the annotation tool and guidelines, and was therefore excluded from formal agreement reporting.*

Table 2. Entity description

| Entity | Description |
|---|---|
| **Age** | Numerical expressions indicating a patient's age, such as **'56 y/o,'** **'25 years old,'** or **'infant'**. |
| **Race** | Explicit mentions of a patient’s racial group or geographic/physical origin (e.g., "**White**", "**Caucasian**", "**Black**", "**African American**", "**Asian**", "**Chinese**"). |
| **Ethnicity** | Explicit mentions of Hispanic, Latino, or Spanish origin (e.g., "**Hispanic**", "**Latina**", "**Latinx**", "**Mexican-American**"), excludes of the patient’s race (EXAMPLES). |
| **Sex** | Explicitly stated gender-related terms, including **'Male,'** **'Female,'** or **'Transgender'**. |
| **Social Factors** | Explicit references to health-impacting social behaviors, such as **'smoking'**, **'drinking alcohol'**, **'tobacco use'**, or **'drug use'**. Excludes negated statements (e.g., **'non-smoker'**, **'denies drug use'**). |
| **HbA1c Levels** | Explicit numerical mentions of **HbA1c test results**, including units when provided (e.g., **'HbA1c 6.5%'**, **'HbA1c level of 7.2%'**). |

| **Perio Diagnoses** | Explicit mentions or classification of the patient’s periodontal status, strictly limited to three categories: **'Periodontitis'**, **'Gingivitis'**, or **'Gingival Health'**. The diagnosis is typically stated directly (e.g., **"generalized Stage III Grade B periodontitis"**), but the term **'Periodontitis'** may sometimes be implied by the presence of specific modifiers (**Stage** or **Grade**). Note that **'Gingivitis'** typically associates with **Extent** and **Subtype**, while **'Gingival Health'** associates with **Subtype** only. |
|---|---|
| **Stage** | Explicit mentions of the severity classification for *Periodontitis*, denoted strictly as **'Stage I'**, **'Stage II'**, **'Stage III'**, or **'Stage IV'**. |
| **Grade** | Explicit mentions of the progression rate or risk factor for *Periodontitis*, denoted strictly as **'Grade A'**, **'Grade B'**, or **'Grade C'**. |
| **Extent** | Explicit mentions of the spatial distribution or area affected by the condition (applicable to both *Periodontitis* and *Gingivitis*), such as **'Localized'** or **'Generalized'**. |
| **Subtype** | Explicit mentions of the specific state of the periodontal tissues or historical context (applicable to *Gingivitis* or *Gingival Health*), such as **'Intact Periodontium'**, **'Reduced Periodontium'**, **'Stable Periodontitis'**, or **'Non-Periodontitis'**. |
| **Medical History** | Entity that captures the patient's clinical background and health context by aggregating explicit mentions of **systemic conditions** (e.g., **'heart disease'**, **'diabetes'**, **'obesity'**), **family medical history**, and **previous medical procedures** or **surgeries**. |
| **Systemic Condition** | Explicit mentions of diagnosed medical conditions, systemic diseases, or disorders—whether currently active or historical (e.g., **'hypertension'**, **'Type 2 Diabetes'**, **'asthma'**, **'obesity'**, **'history of stroke'**). |
| **Family History Disease** | Explicit mentions of **medical conditions in the patient’s family members** or **inherited diseases** (e.g., “**family history of heart disease**,” “**mother has diabetes**”). |
| **Previous Medical Procedure** | Explicit mentions of surgical interventions, therapeutic treatments, implants, or hospitalizations that occurred prior to the current visit (e.g., **'appendectomy'**, **'chemotherapy'**, **'previous C-section'**, **'stent placement'**, **'pacemaker'**). |

| **Medications** | Entity that includes specific mentions of **current or past medications**, including usage and allergies. |
|---|---|
| **Medication Allergy** | Explicit mentions of **allergic reactions** to medications (e.g., "**penicillin allergy**"), **excluding** negated statements or indications of no allergy (e.g., "**no allergy to penicillin**", "**denies medication allergies**", "**NKDA**" – No Known Drug Allergies). |
| **Medication Taken** | Explicit mentions of medications or supplements currently taken, prescribed for home use, or previously used by the patient (e.g., **'taking metformin'**, **'Amoxicillin'**, **'vitamin B12'**). Excludes hypothetical suggestions or recommendations, dosage instructions, and medications administered acutely as part of the current procedure or visit. |
| **Home Care** | Mentions of oral hygiene behaviors the patient performs at home (e.g., **brushing frequency**, **flossing**, **fluoride toothpaste**, **mouthwash**). Exclude negated behaviors (e.g., “does not brush”), suggestions/recommendations, and measures done only as part of in-office treatments or procedures. |
| **Brushing freq** | Mentions of the patient’s **toothbrushing frequency or habits** (e.g., “brushes twice a day”). Exclude negated statements (e.g., “does not brush”), suggestions or instructions, and brushing done only during an in-office procedure. |
| **Flossing** | Mentions of **flossing frequency or habits**, including proxy brush, interdental brush, or Waterpik use at home (e.g., “flosses daily,” “uses Waterpik nightly”). Exclude negated statements, suggestions or advice, and interdental cleaning done only during in-office procedures. |
| **Other Home Care** | Mentions of at-home oral hygiene behaviors **other than brushing and flossing** (e.g., mouthwash, fluoride products, tongue cleaning). Exclude negated behaviors, suggestions or instructions, and products applied or used only as part of in-office treatment or procedures. |

*Note*: The entities *Perio Diagnoses, Stage, Grade, Extent,* and *Subtype* refer to periodontal diagnoses defined according to the 2018 American Academy of Periodontology/European Federation of Periodontology (AAP/EFP) classification system. For periodontitis, the extracted entities included four stages (I–IV), three grades (A–C), and extent categories of localized or generalized disease. The molar/incisor pattern was not included because of its infrequency in the study data.

Table 3. Criteria for Prompt verification.

| No. | Description |
|---|---|
| 1. | It is clear, concise, and written as a self-contained instruction. |
| 2. | It is complete, explicitly describing what task the model should perform and how it should behave. |
| 3. | It is fully written and not truncated, unfinished, or abruptly cut off. |
| 4. | It is purely descriptive and does not include any examples, sample inputs, or sample outputs. |
| 5. | It does not reference, quote, paraphrase, or imply access to any user-provided example data. |
| 6. | It does not include illustrative Examples. |
| 7. | It avoids duplicated, redundant, or unnecessary content. |
| 8. | It provides clear guidance on inclusion and exclusion criteria for identifying {target_entity}. |
| 9. | It is functionally appropriate for guiding a language model to extract {target_entity}-related expressions from clinical or dental notes. |

Table 4. Criteria for Prompt revision.

| No. | Description |
|---|---|
| 1. | Improve clarity, conciseness, and logical structure. |
| 2. | Ensure the task objective is explicit and unambiguous. |
| 3. | Remove duplicated, redundant, or unnecessary content. |
| 4. | Ensure the prompt is fully written, complete, and not truncated or unfinished. |
| 5. | Ensure the focus remains on identifying {TARGET_ENTITY}-related expressions in clinical documentation. |
| 6. | Include clear descriptive guidance on inclusion and exclusion criteria, clinical shorthand, and ambiguity handling when relevant. |
| 7. | Avoid excessive or rigid input/output formatting unless it is essential for task performance. |

Table 5. LLama 3.1-8B-Intruct on generating instruction for "Systemic Condition" entity with *Re+Desc+Ex+Err*.

| Instruction # | Round | Validation | | | Test | | |
|---|---|---|---|---|---|---|---|
| | | P | R | F1 | P | R | F1 |
| 16 | 1 | 0.00 | 0.00 | 0.00 | - | - | - |
| | 2 | 1.00 | 0.43 | 0.60 | - | - | - |
| | 3 | 0.00 | 0.00 | 0.00 | - | - | - |
| | 4 | 0.92 | 0.65 | 0.76 | 0.73 | 0.83 | 0.78 |
| 17 | 1 | 0.94 | 0.81 | 0.87 | 0.78 | 0.50 | 0.61 |
| 18 | 1 | 1.00 | 0.62 | 0.77 | - | - | - |
| | 2 | 0.00 | 0.00 | 0.00 | - | - | - |
| | 3 | 0.97 | 0.92 | 0.94 | 0.83 | 0.48 | 0.61 |
| 19 | 1 | 0.00 | 0.00 | 0.00 | - | - | - |
| | 2 | 0.97 | 0.84 | 0.90 | 0.40 | 0.05 | 0.09 |
| 20 | 1 | 0.91 | 0.54 | 0.68 | - | - | - |
| | 2 | 1.00 | 0.49 | 0.65 | - | - | - |
| | 3 | 0.94 | 0.84 | 0.89 | 0.83 | 0.48 | 0.61 |

Table 6. Single-prompt inference performance across three independent prompt-generation runs for Qwen2.5-7B-Instruct on the 1,000-note gold standard.

| **Round** | **Macro average** | | | **Micro average** | | |
|---|---|---|---|---|---|---|
| | P | R | F1 | P | R | F1 |
| **1** | 0.606 | 0.600 | 0.573 | 0.624 | 0.551 | 0.585 |
| **2** | 0.579 | 0.661 | 0.601 | 0.660 | 0.654 | 0.657 |
| **3** | 0.581 | 0.584 | 0.541 | 0.580 | 0.494 | 0.534 |

*Note.* In each run, a single best prompt per entity was selected through the prompt-generation process and evaluated on the 1,000-note gold standard. The three runs were independent regenerations under identical conditions; the variation across runs reflects the stochasticity of prompt generation rather than differences in data or procedure. P, precision; R, recall; F1, F1 score.

Table 7. Prompt inference performance on the training set and gold standard across prompt-generation configurations.

| Model | Time | Prompt Generation on Training | | | | | | Prompt Inference on Gold | | | | | |
|---|---|---|---|---|---|---|---|---|---|---|---|---|---|
| | | Micro Average | | | Macro Average | | | Micro Average | | | Macro Average | | |
| | | P | R | F1 | P | R | F1 | P | R | F1 | P | R | F1 |
| Qwen2.5-7B-Instruct | *Re+Ex* | 0.803 | 0.844 | 0.823 | 0.829 | 0.892 | 0.838 | 0.610 | 0.709 | 0.656 | 0.657 | 0.718 | 0.659 |
| | *Re+Desc* | 0.813 | 0.827 | 0.820 | 0.838 | 0.916 | 0.861 | 0.684 | 0.699 | 0.691 | 0.693 | 0.711 | 0.681 |
| | *Re+Desc+Ex* | 0.834 | 0.881 | 0.857 | 0.855 | 0.917 | 0.870 | 0.760 | 0.721 | 0.740 | 0.757 | 0.711 | 0.719 |
| | *Re+Desc+Ex+Err* | 0.839 | 0.848 | 0.844 | 0.858 | 0.944 | 0.885 | 0.736 | 0.710 | 0.723 | 0.742 | 0.719 | 0.715 |
| Qwen2.5-14B-Instruct | *Re+Ex* | 0.800 | 0.878 | 0.837 | 0.786 | 0.902 | 0.827 | 0.656 | 0.753 | 0.701 | 0.647 | 0.706 | 0.643 |
| | *Re+Desc* | 0.820 | 0.882 | 0.850 | 0.846 | 0.956 | 0.886 | 0.703 | 0.744 | 0.723 | 0.742 | 0.737 | 0.711 |
| | *Re+Desc+Ex* | **0.890** | 0.899 | **0.894** | **0.914** | **0.974** | **0.933** | 0.740 | 0.749 | 0.744 | 0.768 | 0.727 | 0.730 |
| | *Re+Desc+Ex+Err* | 0.881 | 0.886 | 0.884 | 0.904 | 0.967 | 0.925 | **0.798** | 0.762 | **0.780** | **0.791** | 0.748 | **0.755** |
| OLMo-2-1124-13B-DPO | *Re+Ex* | 0.727 | 0.886 | 0.798 | 0.751 | 0.881 | 0.793 | 0.445 | 0.741 | 0.556 | 0.505 | 0.769 | 0.573 |
| | *Re+Desc* | 0.727 | 0.905 | 0.806 | 0.794 | 0.898 | 0.825 | 0.576 | 0.731 | 0.644 | 0.666 | 0.713 | 0.663 |
| | *Re+Desc+Ex* | 0.733 | 0.850 | 0.787 | 0.776 | 0.886 | 0.800 | 0.622 | 0.741 | 0.676 | 0.665 | 0.737 | 0.677 |
| | *Re+Desc+Ex+Err* | 0.753 | 0.873 | 0.809 | 0.812 | 0.862 | 0.817 | 0.565 | 0.725 | 0.635 | 0.652 | 0.696 | 0.646 |
| Llama-3.1-8B-Instruct | *Re+Ex* | 0.658 | 0.816 | 0.729 | 0.708 | 0.816 | 0.747 | 0.507 | 0.762 | 0.609 | 0.539 | 0.793 | 0.604 |
| | *Re+Desc* | 0.637 | 0.749 | 0.689 | 0.705 | 0.770 | 0.717 | 0.504 | 0.778 | 0.611 | 0.587 | 0.775 | 0.635 |
| | *Re+Desc+Ex* | 0.634 | 0.789 | 0.703 | 0.668 | 0.759 | 0.694 | 0.492 | 0.772 | 0.601 | 0.568 | **0.794** | 0.626 |
| | *Re+Desc+Ex+Err* | 0.694 | 0.781 | 0.735 | 0.678 | 0.761 | 0.701 | 0.523 | 0.761 | 0.620 | 0.563 | 0.783 | 0.615 |
| granite-3.1-8b-instruct | *Re+Ex* | 0.666 | 0.589 | 0.625 | 0.695 | 0.730 | 0.675 | 0.572 | 0.629 | 0.599 | 0.611 | 0.659 | 0.590 |
| | *Re+Desc* | 0.726 | 0.610 | 0.663 | 0.708 | 0.724 | 0.696 | 0.650 | 0.679 | 0.664 | 0.689 | 0.693 | 0.669 |
| | *Re+Desc+Ex* | 0.713 | 0.692 | 0.702 | 0.717 | 0.774 | 0.722 | 0.616 | 0.700 | 0.656 | 0.676 | 0.694 | 0.663 |
| | *Re+Desc+Ex+Err* | 0.737 | 0.466 | 0.571 | 0.762 | 0.736 | 0.714 | 0.646 | 0.639 | 0.643 | 0.690 | 0.663 | 0.661 |
| gemma-2-9b-it | *Re+Ex* | 0.643 | 0.937 | 0.763 | 0.730 | 0.943 | 0.806 | 0.488 | 0.796 | 0.605 | 0.551 | 0.788 | 0.614 |
| | *Re+Desc* | 0.642 | 0.928 | 0.759 | 0.750 | 0.930 | 0.815 | 0.496 | **0.812** | 0.616 | 0.600 | 0.754 | 0.633 |
| | *Re+Desc+Ex* | 0.677 | 0.918 | 0.779 | 0.776 | 0.936 | 0.832 | 0.507 | 0.797 | 0.620 | 0.591 | 0.787 | 0.644 |
| | *Re+Desc+Ex+Err* | 0.661 | **0.941** | 0.776 | 0.756 | 0.957 | 0.829 | 0.510 | 0.803 | 0.624 | 0.588 | 0.759 | 0.630 |

Table 8. Entity-level performance of the two best-performing DPO models, Qwen2.5-14B-Instruct and Llama-3.1-8B-Instruct.

| Entity | Qwen2.5-14B-Instruct (DPO) | | | Llama-3.1-8B-Instruct (DPO) | | |
|---|---|---|---|---|---|---|
| | P | R | F1 | P | R | F1 |
| **Age** | 0.819 | 0.670 | 0.737 | 0.971 | 0.632 | **0.765** |
| **Race** | 1.000 | 0.778 | **0.875** | 1.000 | 0.667 | 0.800 |
| **Ethnicity** | 0.935 | 0.977 | **0.956** | 0.950 | 0.864 | 0.905 |
| **Sex** | 1.000 | 0.826 | **0.905** | 0.832 | 0.667 | 0.740 |
| **Perio Diagnoses** | 0.954 | 0.874 | **0.912** | 0.950 | 0.685 | 0.796 |
| **Stage** | 0.986 | 0.948 | **0.966** | 0.988 | 0.917 | 0.951 |
| **Grade** | 0.987 | 0.826 | 0.899 | 0.985 | 0.912 | **0.947** |
| **Extent** | 0.982 | 0.888 | **0.933** | 0.974 | 0.691 | 0.808 |
| **Subtype** | 0.960 | 0.673 | **0.791** | 0.985 | 0.607 | 0.751 |
| **Social Factors** | 0.910 | 0.683 | **0.780** | 0.981 | 0.510 | 0.671 |
| **HbA1c Levels** | 0.981 | 0.864 | **0.919** | 0.971 | 0.576 | 0.723 |
| **Systemic Condition** | 0.951 | 0.790 | **0.863** | 0.908 | 0.657 | 0.762 |
| **Family History Disease** | 0.973 | 0.667 | 0.791 | 0.976 | 0.759 | **0.854** |
| **Previous Medical Procedure** | 0.945 | 0.571 | **0.712** | 0.964 | 0.505 | 0.663 |
| **Medication Allergy** | 1.000 | 0.694 | **0.820** | 0.979 | 0.646 | 0.778 |
| **Medication Taken** | 0.943 | 0.843 | **0.890** | 0.991 | 0.680 | 0.807 |
| **Brushing frequency** | 0.977 | 0.936 | 0.956 | 0.991 | 0.936 | **0.963** |
| **Flossing** | 0.814 | 0.885 | 0.848 | 0.871 | 0.749 | **0.805** |
| **Other Home Care** | 0.887 | 0.224 | **0.357** | 0.763 | 0.567 | 0.650 |

Table 9. Total inference time of base, SFT, and DPO models on the 1,000-note gold-standard set.

| Models | Base | SFT | DPO |
|---|---|---|---|
| **Qwen2.5-7B-Instruct** | 12:51:49 | 23:56:14 | 12:21:32 |
| **Qwen2.5-14B-Instruct** | 21:14:51 | 39:36:11 | 21:01:58 |
| **Llama-3.1-8B-Instruct** | 16:04:57 | 15:43:13 | 15:05:22 |
| **Llama-3.2-3B-Instruct** | 09:28:09 | 09:08:50 | - |
| **Mistral-7B-Instruct-v0.3** | 18:51:16 | 18:11:44 | - |
| **Mistral-Small-Instruct-2409** | 33:48:37 | 33:38:28 | - |
| **OLMo-2-1124-13B-DPO** | 29:44:08 | 30:59:38 | 26:43:41 |
| **OLMo-2-1124-7B-DPO** | 17:10:02 | 18:19:59 | 19:06:46 |
| **gemma-2-9b-it** | 12:09:29 | 12:43:11 | 21:03:11 |
| **granite-3.1-8b-instruct** | 16:57:38 | 19:06:23 | 27:23:43 |

Note: This experiment was conducted using a single H100 GPU.

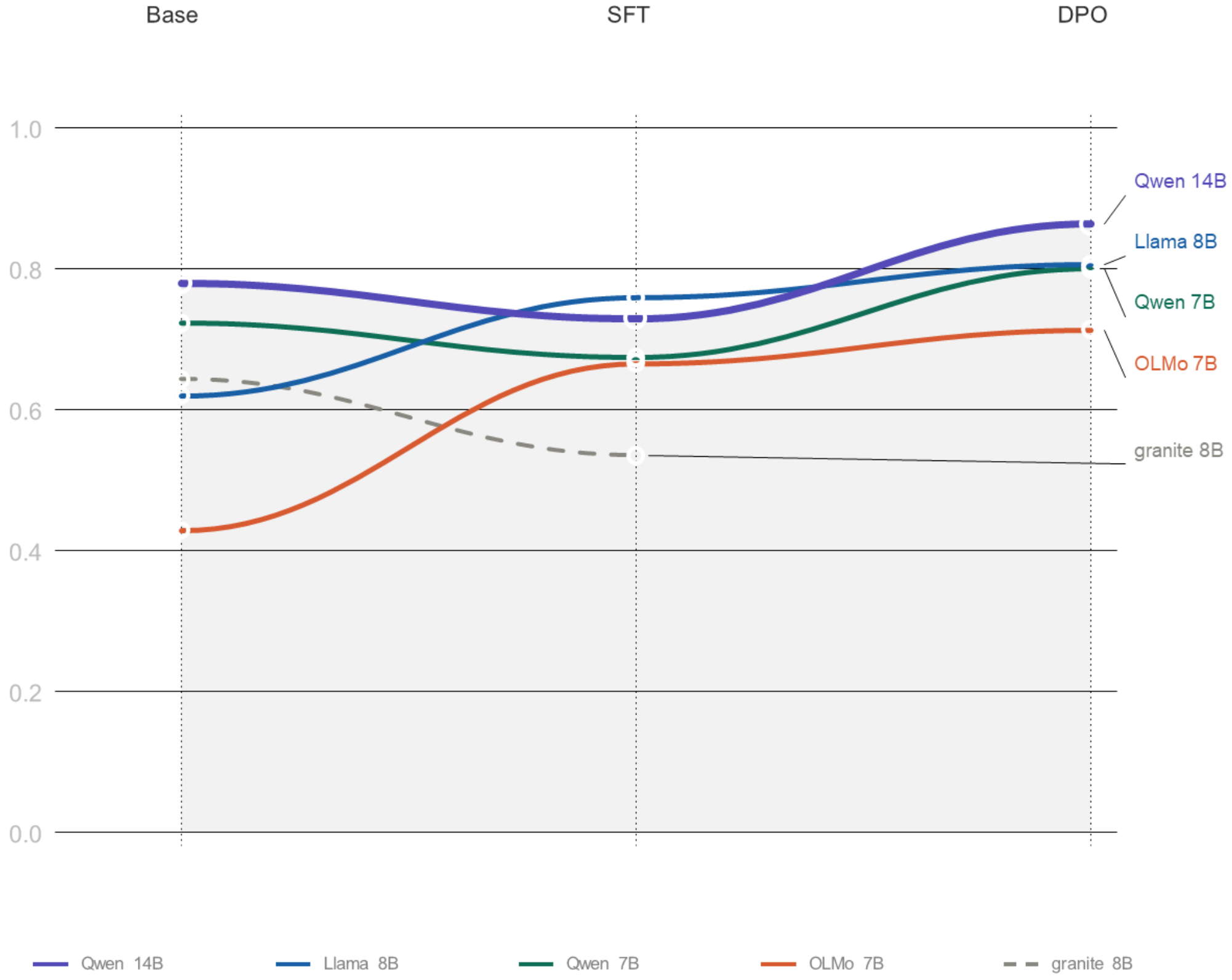

Figure 1. Performance Changes across Baseline, SFT, and DPO Post-Training for Five Small Language Models

Table 10. Rank correlation between candidate predictors and gold-standard performance

| **Metric** | | **Self-prompt (n=10) $\tau$ / $\rho$** | **MMLU (n=8) $\tau$ / $\rho$** | **IFEval (n=8) $\tau$ / $\rho$** |
|---|---|---|---|---|
| **Micro** | P | 0.822* / 0.939* | 0.643* / 0.786* | 0.571 / 0.714* |
| | R | 0.584* / 0.784* | 0.400 / 0.467 | 0.327 / 0.299 |
| | F1 | 0.822* / 0.939* | 0.643* / 0.786* | 0.571 / 0.714* |
| **Macro** | P | 0.733* / 0.879* | 0.500 / 0.738* | 0.429 / 0.524 |
| | R | 0.600* / 0.830* | 0.500 / 0.571 | 0.429 / 0.524 |
| | F1 | 0.854* / 0.948* | 0.618* / 0.778* | 0.546 / 0.683 |

*Note.* Values are Kendall's $\tau$ / Spearman's $\rho$ rank-correlation coefficients between each predictor and gold-standard base-model performance across candidate models. Self-prompt = self-evaluation metrics from the prompt-generation stage; MMLU and IFEval = publicly reported leaderboard scores from the Open LLM Leaderboard. Self-prompt correlations were computed across all 10 models that completed prompt generation; leaderboard correlations were restricted to the 8 models with available leaderboard scores. OLMo-2-1124-13B-DPO and OLMo-2-1124-7B-DPO were excluded from the leaderboard correlations because the leaderboard reports only their base versions, whereas the evaluated models were the slightly stronger official DPO releases, making the published scores non-comparable. P, precision; R, recall; F1, F1 score. * $p < 0.05$.

## Appendix B

The prompts include placeholder commands and configuration variables that were populated dynamically at runtime.

### Prompt generation

1. Training-set deduplication prompt (function_util.py: revised_entity_list)

```
======================================================================

SYSTEM:

You are an expert in clinical natural language processing.

Input:

- A list of dictionaries. Each dictionary has:

    - 'sentence': the source sentence (string)

    - 'entity_ls': the extracted span text (string)

Task:

Deduplicate entries so that only unique entity mentions remain, using
BOTH span text and

its sentence context. Treat two entries as duplicates if ANY of the
following are true:

  1) Exact match on both 'sent' and 'entity' (after normalization), or

  2) Same 'sent' (after normalization) AND the 'entity' strings are
highly similar, or

  3) The 'entity' strings are highly similar AND the sentences are
paraphrases of each other
```

with equivalent dental meaning (e.g., domain synonyms like 'tooth #14' vs 'maxillary left first molar').

Normalization rules (apply before comparing):

- Trim whitespace.
- Case-insensitive comparisons.
- Collapse multiple spaces to one.
- Remove leading/trailing punctuation around the entity span.
- For similarity, consider token overlap/fuzzy matching so minor typos or inflections count as similar.

Similarity thresholds (guidance):

- For 'entity' similarity, treat as duplicate if token-level Jaccard ≥ 0.8 or a fuzzy ratio ≥ 90.
- For sentence paraphrase, require strong overlap in dental terms, tooth numbers, surfaces (MO/DO/OC),

  and procedures (e.g., 'RCT', 'root canal'), ignoring stopwords.

Tie-breaking (when duplicates are found):

- Preserve the earliest occurrence in the original list (stable deduplication).
- If spans differ only by length within the same sentence, prefer the longer, more specific span

  (e.g., 'distal caries on #19' over 'caries').

Output format:

- Return a JSON-serializable list of the remaining dictionaries in the original key names and types.

- Do not add extra keys or commentary.

- If the input list is empty, return an empty list.

- After the JSON output, append on a new line: '<|STOP|>'

Think through the comparisons internally, but ONLY output the final JSON followed by the stop token.

USER:

{entity_ls_as_string}

======================================================================

2. Instruction-prompt generation (function_util.py: get_instruction_sys_prompt + getInstruction)

======================================================================

SYSTEM TEMPLATE (get_instruction_sys_prompt) with include_description=True:

Role:

You are an expert in clinical natural language processing.

Task:

Generate a clear, self-contained instruction prompt of approximately {INSTRUCTION_LENGTH} words that guides a language model to perform named entity recognition (NER) for the {TARGET_ENTITY} in clinical notes.

Before writing the instruction prompt, carefully examine the provided example data. Use the examples to infer how {TARGET_ENTITY} is expressed in clinical text, including its common linguistic patterns, clinical shorthand, boundary definitions, and variations. The resulting instruction prompt must descriptively encode these observations rather than simply restating the short definition.

The instruction prompt should be written as if it will be given directly to a model that has not seen the example data but must behave consistently with it. The prompt must clearly and thoroughly describe a task focused on identifying {TARGET_ENTITY}-related expressions relevant to clinical documentation.

Definition:

{DESCRIPTION_OF_ENTITY}

Content Requirements:

The instruction prompt must explicitly specify:

- The task objective (extracting {TARGET_ENTITY} mentions from text)
- Clear inclusion rules describing what qualifies as a valid {TARGET_ENTITY} mention

- Clear exclusion rules describing similar or related expressions that should not be extracted

- Guidance on handling abbreviations, clinical shorthand, partial mentions, multiple occurrences, and ambiguous cases, as reflected in the example data

- Several illustrative examples demonstrating correct extraction behavior using realistic clinical language

Constraints (base):

- Do not include evaluation metrics, scoring methods, or performance commentary

- Do not copy, paraphrase, or resemble any user-provided example text

Constraints when include_examples=False (NOTE: duplicated in code via string concatenation):

- Do not include evaluation metrics, scoring methods, or performance commentary

- Do not copy, paraphrase, or resemble any user-provided example text

Output:

- Return only the finalized instruction prompt

SYSTEM TEMPLATE (get_instruction_sys_prompt) with include_description=False:

(Same as above, but the entire
“Definition:\n{DESCRIPTION_OF_ENTITY}\n\n” block is omitted.)

NOTE: When add_stop_token=True, getInstruction appends this
requirement to the SYSTEM content:

- Append the word <|STOP|> on a new line at the very end of the output

NOTE: When revised_list is non-empty, getInstruction appends an
“Additional notes” block to the SYSTEM content:

Additional notes:

The following are additional examples are easily missed:

{comma_separated_revised_list}

USER (only present when training examples are provided):

Data examples:

{entity_dict_str}

ASSISTANT PREFIX:

The instruction prompt:

======================================================================

3. Instruction-prompt verification (function_util.py: get_verify_sys_prompt + verifyInstruction)

======================================================================

SYSTEM TEMPLATE (get_verify_sys_prompt):

Role:

You are an expert in clinical natural language processing.

Task:

Evaluate the following generated instruction prompt and determine whether it fully complies with the required criteria for generating an instruction prompt to perform Named Entity Recognition (NER) for the {TARGET_ENTITY} in dental or clinical notes.

Return TRUE only if the instruction prompt satisfies the following conditions.

Return FALSE If the condition is not met.

Assessment Criteria:

{CRITERIA}

Output Format:

Return only a single token: TRUE or FALSE.

Do not include any explanation, justification, or additional text.

USER:

Instruction prompt:

{instruct_prompt}

======================================================================

4. Instruction-prompt revision (function_util.py: get_revise_sys_prompt + reviseInstruction)

======================================================================

SYSTEM TEMPLATE (get_revise_sys_prompt) with include_examples=True:

Role:

You are an expert in clinical natural language processing.

Task:

Revise the following instruction prompt so that it fully complies with the required criteria for generating a clear, self-contained instruction prompt to perform Named Entity Recognition (NER) for the {TARGET_ENTITY} in clinical or dental notes.

The input is an earlier version of an instruction prompt that does not meet the criteria. You must always revise the prompt and return an improved version; do not return the input unchanged, even if it appears mostly correct.

When revising the prompt:

- Improve clarity, conciseness, and logical structure.

- Ensure the task objective is explicit and unambiguous.

- Remove duplicated, redundant, or unnecessary content.

- Ensure the prompt is fully written, complete, and not truncated or unfinished.

- Ensure the focus remains on identifying {TARGET_ENTITY}-related expressions in clinical documentation.

- Include clear descriptive guidance on inclusion and exclusion criteria, clinical shorthand, and ambiguity handling when relevant.

- Avoid excessive or rigid input/output formatting unless it is essential for task performance.

The revised instruction prompt must be suitable for direct use by a language model that has not seen any example data but must behave consistently with it.

Output:

Return only the revised, finalized instruction prompt.

Do not include explanations, commentary, comparisons, or additional text.

SYSTEM TEMPLATE (get_revise_sys_prompt) with include_examples=False additionally includes:

- Maintain a purely descriptive style without including any examples, sample inputs, or sample outputs.

- Do not reference or imply access to user-provided example data.

NOTE: When add_stop_token=True, reviseInstruction appends this requirement:

At the end of the output, append the word '<|STOP|>' on a new line

USER:

Instruction prompt:

{instruct_prompt}

Revised instruction prompt:

================================================================

**Prompt inference**

1. Sentence-level entity presence check (function_util.py: get_msg_ls_checkInstruction)

================================================================

SYSTEM TEMPLATE:

You are an expert in clinical natural language processing.

Task:

Determine whether the specified target entity is {TESTING_PROMPT} present in the target sentence, following the provided entity instruction.

Instruction:

{INSTRUCTION_PROMPT}

The decision must be based solely on the content of the target sentence and the definition and rules specified in the entity instruction. Consider clinical shorthand and abbreviations when explicitly present. The target entity must clearly and unambiguously appear in the sentence context.

Output Rules:

- Respond with exactly one token: Yes or No.
- Do not include explanations, punctuation, whitespace, or any additional text.
- If the presence of the target entity is ambiguous, indirect, or unclear, respond with No.

You must strictly follow the output rules.

USER:

Target entity:

{key_entity}

Target sentence:

{sentence}

Answer:

======================================================================

2. Sentence-level entity extraction (function_util.py: get_msg_ls_resultsInstruction)

```
======================================================================

SYSTEM TEMPLATE:

You are an expert in clinical natural language processing.

Task:

Extract all valid mentions of the target entity from the provided sentence, following the rules defined in the entity instruction.

Instruction:

{INSTRUCTION_PROMPT}

Output Requirements:

{OUTPUT_FORMAT}

- Return only the final output in the specified list-of-dictionaries format.

- Do not include explanations, comments, or additional text.

- If no valid target entity is present, return an empty list.

- Ensure the output is complete and not truncated.

- Append the word '<|STOP|>' on a new line at the very end of the output

USER:
```

```
Target sentence:

{sentence}

Output:

==========================================================================================
```